%% file: main.tex
\documentclass[conf]{new-aiaa}
\usepackage[utf8]{inputenc}

\usepackage{graphicx}
\usepackage{amsmath}
\DeclareMathOperator*{\argmax}{arg\,max}

\usepackage[version=4]{mhchem}
\usepackage{siunitx}
\usepackage{longtable,tabularx}
\usepackage{booktabs}
\usepackage{subfig}
\title{Reward Function Optimization of a Deep Reinforcement Learning Collision Avoidance System}

\author{Cooper Cone\footnote{Military Fellow, MIT Lincoln Laboratory}, Michael Owen\footnote{Technical Staff, Surveillance Systems, MIT Lincoln Laboratory, AIAA Senior Member}, Luis Alvarez\textsuperscript{†}, and Marc Brittain\textsuperscript{†}}
\affil{MIT Lincoln Laboratory, Lexington, MA, 02421}

\begin{document}

\maketitle

\input{abstract}

\let\thefootnote\relax\footnotetext{DISTRIBUTION STATEMENT A. Approved for public release. Distribution is unlimited. This material is based upon work supported by the Dept of Transportation under Air Force Contract No. FA8702-15-D-0001. Any opinions, findings, conclusions or recommendations expressed in this material are those of the author(s) and do not necessarily reflect the views of the Dept of Transportation.This document is derived from work done for the FAA (and possibly others), it is not the direct product of work done for the FAA. The information provided herein may include content supplied by third parties. Although the data and information contained herein has been produced or processed from sources believed to be reliable, the Federal Aviation Administration makes no warranty, expressed or implied, regarding the accuracy, adequacy, completeness, legality, reliability, or usefulness of any information, conclusions or recommendations provided herein. Distribution of the information contained herein does not constitute an endorsement or warranty of the data or information provided herein by the Federal Aviation Administration or the U.S. Department of Transportation. Neither the Federal Aviation Administration nor the U.S. Department of Transportation shall be held liable for any improper or incorrect use of the information contained herein and assumes no responsibility for anyone’s use of the information. The Federal Aviation Administration and U.S. Department of Transportation shall not be liable for any claim for any loss, harm, or other damages arising from access to or use of data information, including without limitation any direct, indirect, incidental, exemplary, special or consequential damages, even if advised of the possibility of such damages. The Federal Aviation Administration shall not be liable for any decision made or action taken, in reliance on the information contained herein. 978-1-5386-6854-2/19/S31.00 2019IEEE}

\input{introduction}

\input{background}

\input{drl_reward_model}

\input{tuning}

\input{results}

\input{conclusion}


\bibliography{references}

\end{document}

%% file: abstract.tex
\begin{abstract}
The proliferation of unmanned aircraft systems (UAS) has caused airspace regulation authorities to examine the interoperability of these aircraft with collision avoidance systems initially designed for large transport category aircraft. Limitations in the currently mandated TCAS led the Federal Aviation Administration to commission the development of a new solution, the Airborne Collision Avoidance System X (ACAS X), designed to enable a collision avoidance capability for multiple aircraft platforms, including UAS. While prior research explored using deep reinforcement learning algorithms (DRL) for collision avoidance, DRL did not perform as well as existing solutions. This work explores the benefits of using a DRL collision avoidance system whose parameters are tuned using a surrogate optimizer. We show the use of a surrogate optimizer leads to DRL approach that can increase safety and operational viability and support future capability development for UAS collision avoidance.
\end{abstract}

%% file: introduction.tex
\section{Introduction}

The Traffic Alert Collision Avoidance System (TCAS) has been an integral part of the increased safety of air transport since it was federally mandated in the 1991 for all passenger carrying aircraft with more than 30 seats flying in U.S. airspace~\cite{lebron1983system, tcas71}. TCAS led to a dramatic reduction in the occurrence of mid air collisions in modern aviation; however the heuristic based approach undertaken in TCAS has made it difficult to adapt the system to the evolving complexity of the National Airspace System (NAS), which includes new cooperative surveillance systems (e.g., ADS-B) and new vehicle entrants. In response, the Federal Aviation Administration (FAA) commissioned the development of a replacement for TCAS. This new system, referred to as the Next Generation Airborne Collision Avoidance System X (ACAS X), which is currently in development at MIT Lincoln Laboratory and John Hopkins Applied Physics Laboratory, is expected to integrate into multiple aircraft platforms and reduce nuisance alerts as well as reduce the risk of Near Mid Air Collisions (NMAC)~\cite{kochenderfer_next-generation_2012}. ACAS X introduced several variants designed to reduce the risk of NMAC for a particular operation, such as commercial aviation (ACAS Xa)~\cite{XaMOPS}, large uncrewed aerial systems (ACAS Xu)~\cite{XuMOPS}, smaller uncrewed aerial vehicles (ACAS sXu)~\cite{sXuMOPS}, and ACAS Xr which is under development for advanced air mobility and helicopter operations. Each variant adds capabilities and design considerations for the operational environment and platforms that will be commonly seen by the ACAS X equipped vehicle. For example, ACAS sXu introduced vehicle to vehicle surveillance to accommodate a future link that sUAS may use to interrogate and coordinate with each other. While, ACAS Xu added Remain Well Clear alerting due to its use in remotely piloted or autonomous UAS. Core to the ACAS X family of collision avoidance systems is the approach of modeling the collision avoidance problem as a Partially Observable Markov Decision Processes (POMDP), which encompasses the state space of interest, allowable actions, the probability of transitioning between states, and reward model dictating the tuning objective.

The POMDP is solved through value iteration, a dynamic programming (DP) algorithm, to calculate a Q-function representing the value gained from taking an action, $a$, from the current state, $s$. In prior collision avoidance solutions the Q-function has been represented as a lookup table that approximates the continuous collision avoidance problem as a set of discretized states and their optimal action~\cite{holland2013}. A shortcoming of the table representation is the memory footprint required to store the combination of all state variables and associated actions. Previous work has attempted to tackle the memory footprint problem of a horizontal collision avoidance logic by representing similar table regions with a single equivalence class and achieving a 50\% reduction in memory footprint with minimal degradation in safety~\cite{julian_deep_2019}. The second approach taken by~\cite{julian_deep_2019} used a representative horizontal logic lookup table as the training objective for a neural network to compress the gigabyte size table into a megabyte size file, but failed to run at the desired 1 Hz cycle required by ACAS X. Thus, the lookup table approach continues to be the solution to maintain the execution constraint in avionics hardware; however the state space variables needed to represent the three dimensional collision avoidance problem would result in an intractable lookup table encompassing terabytes of memory. To overcome the dimensionality curse, previous work has focused on dividing the collision problem into two sub-problems, horizontal and vertical collision avoidance~\cite{alvarez_acas_2019,owen_acas_2019} with separate lookup tables. The independent POMDP solutions produce collision avoidance systems that are found to be safe, operationally suitable, and robust to increased airspace complexity. Nevertheless assuming compute power will increase, representing the POMDP as a neural network may allow an integrated three dimensional logic to reduce alerting complexity and increase safety.

An alternative solution used this approach and solved the POMDP using deep reinforcement learning~\cite{corteguera_airborne_2020}. The multi-layer perceptron (MLP) neural network representation eliminated the need to separate the collision problem into two sub-problems, allowing for a greater level of coordination between the two axes. Although this approach used the POMDP specified by the DP approach~\cite{owen_acas_2019} its performance fell short of the DP approach, partially because the parameters for the reward model were not properly optimized. This work attempts to improve the performance of the model developed by \cite{corteguera_airborne_2020} by introducing a surrogate optimization technique to optimize the parameters of the reward model. The  remainder of the paper is organized as follows.  Section \ref{sec:background} provides background information relevant to this work. Section \ref{sec:drl_reward_model} describes the structure of the reward model. Section \ref{sec:parameter_optimization} applies the surrogate optimizer to the reward model. Section \ref{sec:results} discusses the results from the modified reward model.

%% file: background.tex
\section{Background} \label{sec:background}

\subsection{Model}

In order to increase safety of the airspace, ACAS X uses a sophisticated model of the environment to solve a Markov decision process (MDP), providing a policy that resolves NMAC \cite{russel_artificial_2015, giannakopoulou_exploring_2016, kaelbling_planning_1998}. Formulating an MDP requires an environment state definition, actions the actor is able to implement, a transition function that describes how the state evolves given an action, and a reward model that defines the desirability of the agent's action \cite{garcia_markov_2013}. A perfect actor will always choose actions that lead to optimal reward values. However, we cannot assume that each actor has a perfect model of the current state, so we modify the MDP into a POMDP, which allows the actors to make decisions with a probabilistic model of the current state \cite{spaan_partially_2012}. This modification is useful for aircraft collision avoidance because the ownship, the aircraft which contains the collision avoidance logic, cannot know the exact locations of intruder aircraft (i.e., due to imperfect sensors). The POMDP for aircraft collision avoidance is defined as the tuple of (S, A, T, R, $\Omega$, O, $\gamma$), where \cite{russel_artificial_2015}:

\begin{itemize}
    \item S is the set of all possible states of an encounter.
    
    \item A is the set of actions the ownship can take.
    
    \item T: $p$(S $|$ S, A) is the state transition function, which gives the probability of transitioning to each other state given the current state and the ownship's action.
    
    \item R: S x A $\rightarrow$ $\mathbb{R}$ is the reward function, where for this situation, we use negative rewards to disincentivize the aircraft from taking actions that result in an undesirable state, such as an NMAC.
    
    \item $\Omega$ is the set of all possible observations, where an observation consists of all the data collected from the ownship’s sensor array.
    
    \item O: $p(\Omega$ $|$ S, A) is the observation function, which gives the probability of arriving at an observation given that the ownship took a specific action that transitioned the environment to a specific state.
    
    \item $\gamma$ is the discount factor, weighing the benefit of immediate rewards versus future rewards.
\end{itemize}


In order to make decisions in the environment, the ownship maintains a policy, $\pi$, that determines which action it should take. In an MDP, this policy is a mapping between states and actions. 

\subsection{Dynamic Programming Solution}

In environments with large state and observation spaces, finding an optimal policy for a general POMDP is more difficult than for an MDP. However, if we assume that uncertainties in observations are accounted for during runtime, we can treat our POMDP as an MDP and solve for an optimal policy with Value Iteration, a Dynamic Programming (DP) algorithm \cite{owen_acas_2019, karkus_qmdp-net_2017}.

Value Iteration generates an optimal policy by calculating the expected future reward for transitioning to each state, which is denoted $V(S)$ \cite{bertsekas_dynamic_2012}. We then use Eq.~\eqref{eq:bellman} to update the state values until $V($S$)$ converges \cite{bellman_markovian_1957}.

\begin{equation}
    \label{eq:bellman}
    V(s)_{i+1} = \max\limits_{a}[R(s, a) + \sum\limits_{s'}p(s'|s,a)V_i(s')]
\end{equation}

For each state and action, $(s, a)$, this update calculates both the immediate and expected rewards for taking action $a$. When determining the expected reward, we examine each future state, $s'$, and calculate the probability of transitioning to $s'$ from $s$ after taking action $a$. We then multiply this probability with the expected reward that we will get from this new state. This update process begins with $V(S)_0 = 0$. After $V(S)$ converges, we use Eq.~\eqref{eq:QFunc} to calculate $Q(s,a)$, which represents the expected reward for taking $a$ from $s$.

\begin{equation}
    \label{eq:QFunc}
    Q(s,a) = \sum\limits_{s'}p(s'|s,a)V(s')
\end{equation}

Finally, we generate the optimal policy by maximizing each state's expected reward for each action.

\begin{equation}
    \pi(s) = \max\limits_{a}Q(s,a)
\end{equation}

We represent $Q(S,A)$ as a table, storing the utility for every possible state and action combination. Because the MDP's states represent continuous variables, we must first discretize them into bins. As long as the steps between stored values are small enough, the learned utility function will be close to optimal. Dividing the collision avoidance problem into two sub-problems results in training two lookup tables with value iteration: one that tracks the utility for left and right actions (i.e. horizontal logic) and one that tracks the utility of the climb and descend actions (i.e. vertical logic). 

\subsection{Deep Reinforcement Learning}

An alternative solution to dynamic programming, is an implementation of a Deep Reinforcement Learning (DRL) algorithm known as DQN \cite{mnih_human-level_2015}. This algorithm uses the POMDP formulation from the DP approach, but instead of treating the Q function as two discretized tables, we use a Multi-Layer Perceptron (MLP) neural network.

We train this neural network in a series of episodes, where each episode constitutes a major lesson learned for the final neural network. During each episode, we use a simulation environment to generate aircraft encounters that serve as the algorithm's training data. These encounters produce aircraft states, actions, and rewards associated with the state. In order to maintain consistency during episodes, we store these generated encounters in a replay buffer. Then, for each step in the episode, we sample a mini-batch of aircraft transitions and update the network using gradient descent. The loss function for DQN can then be formulated as

\begin{equation}
    \label{eq:lossFunc}
    L(s,s',a,r|\theta,\theta') = (\max\limits_{a'}Q_{\theta'}(s',a') + r - Q_\theta(s,a))^2.
\end{equation}

Prior research has shown that the loss function in Equation~\eqref{eq:lossFunc} leads the policy to overestimate state-action values. To mitigate this issue, we instead use the double Q-learning technique that replaces the max operation in  Equation~\eqref{eq:lossFunc} with an argmax operator~\cite{van_hasselt_deep_2016}. The double Q-learning loss can then be formulated as

\begin{equation}
    \label{eq:doubleLossFunc}
    L(s,s',a,r|\theta,\theta') = (Q_{\theta'}(s',\argmax\limits_{a'}Q_{\theta}(s',a')) + r - Q_\theta(s,a))^2.
\end{equation}

During training, we calculate emitted rewards using a configurable reward model. This model only depends on the state and tracks four costs: NMAC, Alert, Reversal, and Cease Alert. These costs allow us to prioritize a safe policy, which also considers operational considerations \cite{corteguera_airborne_2020}.

%% file: drl_reward_model.tex
\section{Deep Reinforcement Learning Reward Model} \label{sec:drl_reward_model}

When training the DQN agent, we use a simulation environment that allows us to generate actions based on the current MLP and receive rewards that update its weights. Our MLP has one input layer of 25 nodes (one for each observation variable), seven hidden layers, each with 512 nodes, and an output layer with nine nodes. Each of these output nodes represents one of nine combined actions, which lie along two dimensions: vertical and horizontal. The vertical actions are CLIMB, CLEAR, and DESCEND, while the horizontal actions are LEFT, CLEAR, and RIGHT. A combined action is then created by selecting one horizontal and one vertical action. The environment generates a series of encounters, where each encounter is defined by the starting state and intruder actions. The initial heading, speed, and vertical rate for the intruder and ownship are sampled from a uniform distribution. The relative position of each aircraft is then selected, such that there is an NMAC about 40\% of the time if the ownship does not act.

The sequence of intruder actions is modeled as a Markov chain, where each new action is only dependent on its previous action. This model has two parameters: one defines the average length of an action, while the other defines the average length an aircraft is clear. This Markov chain is then defined such that the intruder aircraft has a high probability of maintaining its previous action and a small probability of transitioning to a different action.

At each update, the ownship and intruder actions are calculated from the MLP and Markov chain, respectively. Their positions are then propagated based on their heading and vertical rate changes. After this update, we give the ownship a reward designed to penalize it for undesirable behavior. Rewards are given for four state results: NMAC, Alert, Reversal, and Cease Alert. If the distance between the intruder and ownship falls below the NMAC threshold, the ownship is given the NMAC cost. To discourage ownship from issuing unnecessary maneuvers, we add a cost to alert, given any time the ownship issues an action. To maintain consistency in the aircraft dynamics, we also penalize reversals, which occur when the ownship switches the direction of a maneuver. Finally, we penalize the ownship for ceasing an alert, to prevent ceasing an alert to avoid the reversal cost by stopping a maneuver in one timestep, then issuing a maneuver in the reversed direction in a future timestep. After a sufficient amount of encounter updates, we finish training and test the model on a separate data set \cite{corteguera_airborne_2020}.

%% file: tuning.tex
\section{Reward Model Parameter Optimization} \label{sec:parameter_optimization}

A surrogate model is used to automatically explore a set of reward parameters that allow the DRL approach to meet the safety and operational suitability objectives. Previous work is leveraged to build the surrogate model and provide data for tuning.

\subsection{Runtime Simulation Environment}

To quantify the performance of each parameter set, we run a series of aircraft simulations using a subset of the Lincoln Laboratory Correlated Encounter Model (LLCEM) encounter set. This set contains 10,000,000 encounters, where each encounter has one intruder aircraft. The maneuvers these intruder aircraft perform were generated from a probabilistic model derived from real radar observations \cite{kochenderfer_correlated_nodate}.

We use a subset of the encounter set, as the DRL approach takes significantly more time to update than the DP approach. The increased update time is in part due to the need to update the states by taking a step in the simulation environment, while the DP approach only explores the states represented in the MDP and queries a value from a table stored in memory. Section \ref{drl_performance} will quantify the DRL update time, while subsequent sections will examine methods to reduce the computational cost.

\subsection{Surrogate Model}

An existing surrogate optimizer tool is leveraged to tune the parameters for the reward model~ \cite{lepird_multi-objective_2015}. Surrogate optimizers are well suited for this problem because the objective function is computationally expensive to evaluate. In addition, traditional optimization techniques such as metaheuristics do not retain complete memory of prior search work and thus require more iterations to converge. This increase in required work is infeasible for the DQN approach due to the extensive simulation required to evaluate the performance of each trained model.

When the surrogate optimizer is applied to the reward model, we optimize three reward parameters: Alert, Reversal, and Cease Alert, where each parameter has the domain [-1, 0]. We assume that the NMAC reward parameter has a value of -1, allowing us to decrease the number of needed search points. We optimize these parameters using three metrics: P(NMAC), P(Alert), and P(Reversal).

This optimization technique selects a point that contains a value for the Alert, Reversal, and Cease Alert costs. We then train a model using the DRL training pipeline and evaluate the performance of that model in our simulation framework using a set metrics that report how often each undesirable state or action occurs. An objective value, discussed in section \ref{Surrogate_Performance_Metrics}, is then used to update the surrogate model, allowing us to select a new point to evaluate.

Before we can use the surrogate optimizer, we must first initialize it with a set of pre-selected points. We generate these points with a three dimensional Latin hypercube, where each dimension represents a reward parameter. 

After evaluating each of the initially sampled points, we construct a surrogate model that approximates the continuous objective function for the entire search space. Each test point is then selected to balance exploration and exploitation. We maintain this balance by calculating the point with the maximum expected improvement, using the surrogate model's objective value mean to explore the search space and the points' variance to exploit known maxima.

\subsection{Runtime Performance} \label{drl_performance}

Before discussing the performance of the DQN and DP algorithms, we will first describe the hardware used for execution and training. Model training and evaluation for both algorithms can be sped-up with 
two different systems within the Lincoln Laboratory Supercomputing Center (LLSC) \cite{reuther_interactive_2018}. The first uses NVIDIA Tesla V100 Graphics Processing Units (GPUs) designed for deep neural network computations, while the second uses high-performance Intel Xeon Platinum 8620 Central Processing Units (CPUs) that include a specialized Single-Instruction-Multiple-Data (SIMD) instruction set.

To compare the runtime updates between the DQN and DP algorithms, we first need to quantify the time taken for each encounter. Because we evaluate these encounters in a simulation framework, we need to consider the overhead from the framework. To do this, we will evaluate a simulation with N encounters and one with a single encounter. We then calculate the time spent in aircraft dynamics with Eq.~\eqref{eq:time}, where all times are reported in seconds:

\begin{equation}
    \label{eq:time}
    T = \frac{T_N - T_1}{N - 1}
\end{equation}

We estimate the performance of each algorithm by running these simulations three times and averaging $T$. To ensure this calculation is reproducible, we evaluate the simulations once before storing the results so that we can try to ensure the necessary memory operations do not result in a page-fault. The results from these simulations for DQN and DP are shown in Table \ref{tab:drl_dp_runtime}.

\begin{table}
    \centering
    \renewcommand{\arraystretch}{1.5}
    \caption{DQN and DP Runtime Performance}
    \begin{tabular}{lrrrr}
        \toprule
        \textbf{Algorithm} & \textbf{Iter 1 (s)} & \textbf{Iter 2 (s)} & \textbf{Iter 3 (s)} & \textbf{Avg (s)} \\
        
        \midrule
        
        DQN & 28.1693 & 27.6316 & 26.9382 & 27.5797 \\
        
        DP & 0.6675 & 0.6761 & 0.6621 & 0.6686 \\

        \bottomrule
        
    \end{tabular}
    \label{tab:drl_dp_runtime}
\end{table}

As we can see from the results, the DQN update cycle is about 41 times slower than the DP update. To improve this, we perform two runtime optimizations on the DQN algorithm. The first is to remove the remain well clear lookahead, which allows the ownship aircraft to predict future conflicts and issue maneuvers earlier. Removing this capability decreases the aircraft's performance but drastically improves the runtime load. This performance loss will not significantly affect parameter optimization, as each trained model will be affected in a similar manner. Second, we utilize the AVX-512 instruction set to vectorize the neural network feed-forward algorithm. The results from these two optimizations are shown in Table \ref{tab:drl_runtime_improvement}

\begin{table}
    \centering
    \renewcommand{\arraystretch}{1.5}
    \caption{DQN Runtime Performance Improvement}
    \begin{tabular}{lrrrrr}
        \toprule
        \textbf{Optimization} & \textbf{Iter 1 (s)} & \textbf{Iter 2 (s)} & \textbf{Iter 3 (s)} & \textbf{Avg (s)} & \textbf{Speedup} \\
        
        \midrule
        
        Lookahead Removal & 13.7242 & 13.7053 & 13.7745 & 13.7347 & 2.0080 \\
        
        Vectorized Instructions & 7.4268 & 7.1785 & 7.6214 & 7.4089 & 3.7225 \\

        \bottomrule
        
    \end{tabular}
    \label{tab:drl_runtime_improvement}
\end{table}

Together, these two performance optimizations allow for a speed-up of 3.722, which means that the DQN update cycle is only about 11 times slower than the DP approach. Even though this improvement is substantial, we still need to evaluate our solutions on a subset of the LLCEM encounter set.

\subsection{Randomness Modification} \label{randomness_modification}

While the surrogate optimizer has been successfully applied to parameter optimization for the DP approach, it may face issues due to the randomization present in the DQN algorithm. Because each model is initialized with random values and its training data is randomly sampled from a replay buffer, we can train two models with identical parameters that finish with different behavior \cite{corteguera_airborne_2020}. Solving this behavior challenge is beyond the scope of this paper, so we must modify the surrogate optimizer to account for the DQN approach's randomness. To accomplish this, we train three different models, evaluate all three, and then choose the model with the best performance. These additional models allow us to be reasonably confident that at least one model is trained well and does not exhibit poor performance.

\subsection{Iteration Completion Time}

Run time requirements should be considered prior to utilizing a surrogate optimizer to understand if a surrogate model is appropriate to solve our problem. To speed up training and evaluation of our model, we streamline the process, by allowing the surrogate optimizer to select a new point for training before the current point has finished evaluation. This modification allows us to begin training a new point while the previous point is still being tested \cite{lepird_multi-objective_2015}. To maximize the system's efficiency, we would like the training process to take the same amount of time as the simulation evaluation. Training each model was found to be most efficiently performed on the LLSC GPUs, taking six hours to complete. As discussed in section \ref{randomness_modification}, we need to generate three different models for each iteration to mitigate the effects of the randomization. Using LLSC, we train in parallel all three models of a given iteration, so that each additional training run adds no time. However, surrogate optimizer restrictions require us to evaluate these three simulations in series, where each simulation can utilize 1000 LLSC CPU cores. Because they are evaluated in series, we would like to select an encounter set such that all three are completed in six hours, to coincide with the training pipeline. If we allow each simulation set two hours to complete, then the training and evaluation processes will avoid stalls. Using the LLCEM encounter set and limiting the evaluation time to two hours, we can complete 50,000 encounters. This encounter set includes a total of 10,000,000 encounters, where most do not contain a nominal NMAC. For example, of the first 50,000 encounters, only 544 issue a nominal NMAC. Because we'd like to ensure the system is safe, we select these 50,000 encounters such that 25,000 have a nominal NMAC and the rest do not. 

\subsection{Performance Metrics} \label{Surrogate_Performance_Metrics}

When evaluating the performance of each model, we will use three metrics: P(NMAC), P(Alert), and P(Reversal). In order to strike a balance between all three, we use the performance of~\cite{owen_acas_2019} as a target, as shown in table \ref{tab:xu_metric_targets}. Each metric is then assigned a score, $S$, which is calculated in Eq.~\eqref{eq:ratio} and \eqref{eq:score}

\begin{table}
    \centering
    \renewcommand{\arraystretch}{1.5}
    \caption{Xu Metric Targets}
    \begin{tabular}{ccc}
        \toprule
        \textbf{P(NMAC)} & \textbf{P(Alert)} & \textbf{P(Reversal)} \\
        
        \midrule
        
        \num{9.8268e-4} & 0.1946 & 0.00290 \\

        \bottomrule
        
    \end{tabular}
    \label{tab:xu_metric_targets}
\end{table}

\begin{equation}
    \label{eq:ratio}
    R = \frac{M}{M_{\text{target}}}
\end{equation}

\begin{equation}
    \label{eq:score}
    S = \frac{R^2}{2} - \frac{1}{2}
\end{equation}

\noindent
where $M$ is the metric value, $M_{\text{target}}$ is the target value from ACAS Xu, and $R$ is the normalized ratio. This function strongly penalizes a metric for failing to meet the target and weakly incentivizes the model to beat the target. This difference allows us to favor policies that get close to all three metrics instead of one that strongly outperforms in a single metric and fails in the other two.

These metric scores are then combined with their weight values, $W$, in the objective function defined by Eq.~\eqref{eq:objectiveFunc}:

\begin{equation}
    \label{eq:objectiveFunc}
    V = W_{\text{P(NMAC)}} \cdot S_{\text{P(NMAC)}} + W_{\text{P(Alert)}} \cdot S_{\text{P(Alert)}} + W_{\text{P(Reversal)}} \cdot S_{\text{P(Reversal)}}
\end{equation}

One of the benefits of the surrogate model is that the metric weight values can be recalculated without requiring any previous simulations to be performed again. The final weight values for P(NMAC), P(Alert), and P(Reversal) were chosen to be 0.05, 0.80, and 0.15, respectively. We chose these values because it is common for many of the trained DRL models to alert close to 100\% in order to decrease P(NMAC). This weighting favors policies with P(Alert) values close to Xu's. These values were continually updated as we gained more information on which policies were being favored.

\subsection{Surrogate Modelling with Instability}

\begin{figure}
    \centering
    \subfloat[Model 1]{\includegraphics[scale=0.4]{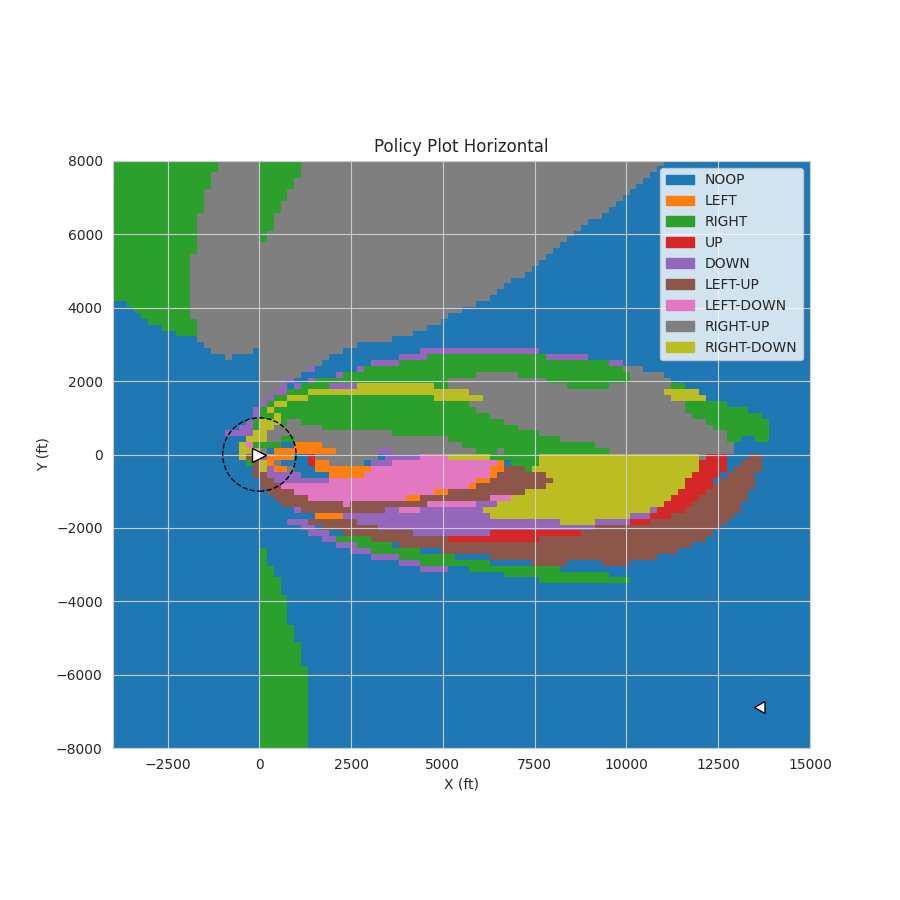}}
    \subfloat[Model 2]{\includegraphics[scale=0.4]{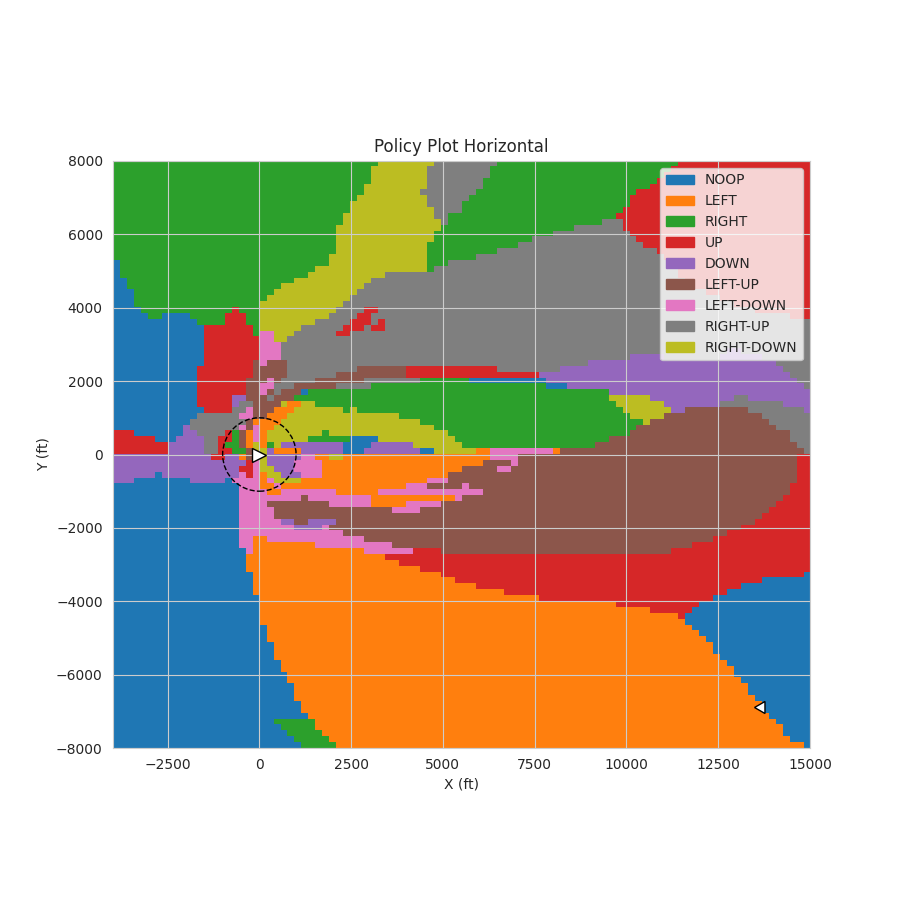}} \\
    \subfloat[Model 3]{\includegraphics[scale=0.4]{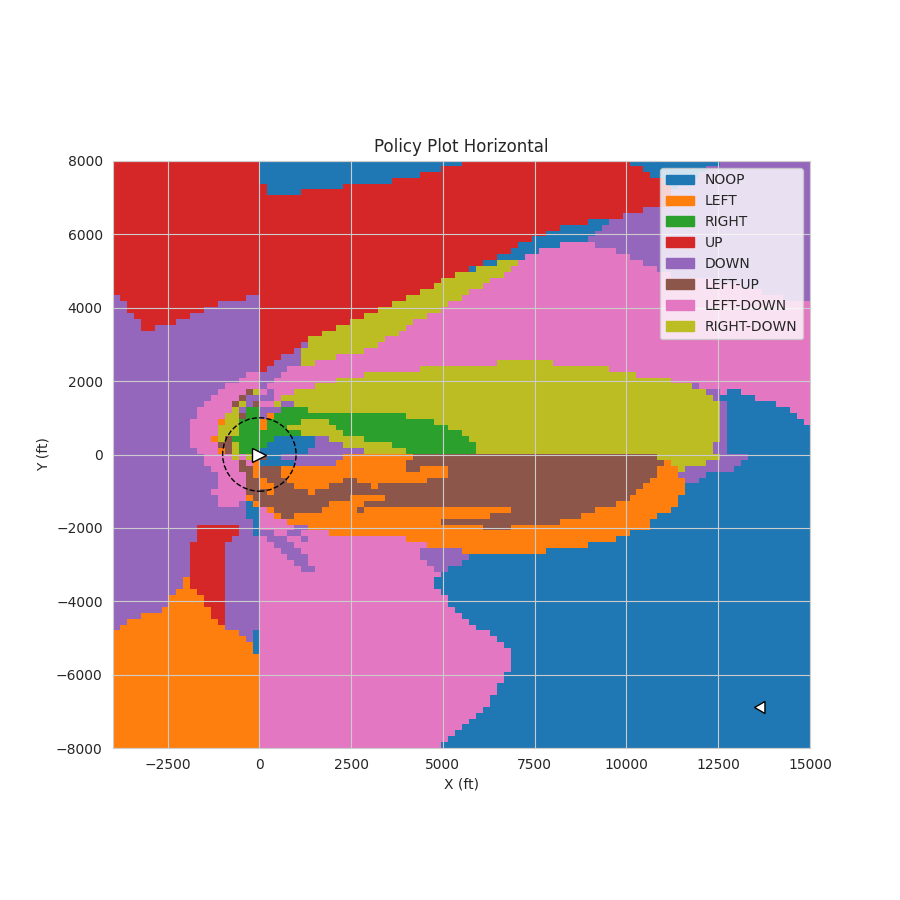}}
    \caption{High Variance Behavior Policy Plots}
    \label{fig:drl-50-plots-high-variance}
\end{figure}

During the tuning process, we discovered that many reward model values did not result in stable policies. For these parameter sets, the three trained models exhibited different behaviors. We analyzed the models through policy plots, which illustrate the actions the ownship aircraft will select in a two-dimensional representation. The policy plots for three models trained from the same reward model parameter set are shown in Fig. \ref{fig:drl-50-plots-high-variance}. The erratic behavior for reward model parameter sets causes issues in the convergence of the surrogate optimizer. We observe the surrogate optimizer would select a high number of test points within these regions because their high metric variance incorrectly implied that there was more exploration to be done. Because of this limitation, we decided to override some of the surrogate optimizer's decisions with hand-selected parameter sets.

These hand-selected sets were chosen with two methods: linear sweeps and Latin-hypercube sampling. When examining the results from the surrogate optimizer, we identified multiple points that exhibited stable and desirable behavior. We then selected $N$ points, whose parameter sets formed a line between two of these known points. An example linear sweep of six points between iterations 50 and 52 is shown in Table \ref{tab:drl-linear-sweep}. To exploit the results around a single point, we used Latin-hypercube sampling to select $N$ points in a region close to the origin point.

\begin{table}
    \centering
    \renewcommand{\arraystretch}{1.5}
    \caption{Linear Sweep Between 50 and 52}
    \begin{tabular}{cccc}
        \toprule
        \textbf{Iteration} & \textbf{Alert} & \textbf{Reversal} & \textbf{Cease Alert} \\

        \midrule
        
        50 & \num{-5.9669e-4} & -0.8324 & -0.1128 \\
        
        52 & \num{-1.3570e-5} & -0.6253 & -0.1018 \\

        \midrule
        
        91 & \num{-5.1339e-4} & -0.8028 & -0.1112 \\
        
        92 & \num{-4.3008e-4} & -0.7732 & -0.1097 \\
        
        93 & \num{-3.4678e-4} & -0.7436 & -0.1081 \\
        
        94 & \num{-2.6348e-4} & -0.7140 & -0.1065 \\
        
        95 & \num{-1.8018e-4} & -0.6844 & -0.1049 \\
        
        96 & \num{-9.6873e-5} & -0.6549 & -0.1034 \\
        
        \bottomrule
        
    \end{tabular}
    \label{tab:drl-linear-sweep}
\end{table}

%% file: results.tex
\section{Results} \label{sec:results}

After tuning the policies manually the surrogate optimizer was utilized to explore 147 iterations. The metrics and policies for the top five iterations are shown in Tables \ref{tab:tuning_metrics} and \ref{tab:tuning_parameters} respectively.

\begin{table}
    \centering
    \renewcommand{\arraystretch}{1.5}
    \caption{Tuning Metrics}
    \begin{tabular}{clllr}
        \toprule
        \textbf{Iteration} & \textbf{pNMAC} & \textbf{pAlert} & \textbf{pReversal} & \textbf{Value} \\

        \midrule
        
        Xu & \num{9.8268e-4} & 0.19460 & 0.00290 & 0.0 \\
        
        Untuned & \num{1.3380e-2} & 0.14950 & 0.03238 & 13.7174 \\

        \midrule
        
        46 & \num{6.0500e-3} & 0.21749 & 0.00780 & 1.4892 \\
        
        52 & \num{7.0343e-3} & 0.32035 & 0.00015 & 1.8643 \\
        
        124 & \num{4.4496e-3} & 0.41754 & 0.00217 & 1.8959 \\
        
        139 & \num{7.1049e-3} & 0.19062 & 0.00938 & 1.9743 \\
        
        138 & \num{7.6419e-3} & 0.20802 & 0.00792 & 2.0268 \\
        
        \bottomrule
        
    \end{tabular}
    \label{tab:tuning_metrics}
\end{table}

\begin{table}
    \centering
    \renewcommand{\arraystretch}{1.5}
    \caption{Tuning Parameters}
    \begin{tabular}{clll}
        \toprule
        \textbf{Iteration} & \textbf{Alert} & \textbf{Reversal} & \textbf{Cease Alert} \\

        \midrule
        
        Untuned & \num{-1.0e-2} & -0.05 & -0.05 \\

        \midrule
        
        46 & \num{-7.0000e-4} & -0.0869 & -0.0130 \\
        
        52 & \num{-1.3570e-5} & -0.6253 & -0.1018 \\
        
        124 & \num{-9.6872e-5} & -0.6549 & -0.1034 \\
        
        139 & \num{-7.1822e-4} & -0.0850 & -0.0138 \\
        
        138 & \num{-6.8891e-4} & -0.0788 & -0.0126 \\
        
        \bottomrule
        
    \end{tabular}
    \label{tab:tuning_parameters}
\end{table}
When examining the top policies, we first quantify the stability of the parameter set. If the reward model parameters are  unstable, then the performance of the trained model is partially attributable to the initialization of the random seed and thus is not a good candidate for the final reward model. Looking at the high variance behavior policy plots in Fig. \ref{fig:drl-50-plots-high-variance}, we can see that one difference between them is the size and location of the regions where the ownship aircraft issues no maneuver. This difference in region size indicates that we can use the variance of the alert metric to help quantify the stability of the reward models. The alert values and variances for each of the top five iterations are shown in Table \ref{tab:drl-top-five-variances}. These results indicate that iterations 52 and 124 are unstable and thus are not good candidates for the final solution. From the remaining iterations, we select iteration 46 because it maintains an acceptable alert level and has the lowest NMAC and reversal rates.

\begin{table}
    \centering
    \renewcommand{\arraystretch}{1.5}
    \caption{Alert Variance for Top Five Models}
    \begin{tabular}{cllll}
        \toprule
        \textbf{Iteration} & \textbf{Model 1} & \textbf{Model 2} & \textbf{Model 3} & \textbf{Variance} \\
        \midrule
        
        46 & 0.2709 & 0.2175 & 0.1941 & 0.0010 \\
        
        52 & 0.1948 & 0.9223 & 0.3203 & 0.1008 \\
        
        124 & 0.4175 & 0.9999 & 0.9999 & 0.0754 \\
        
        139 & 0.1669 & 0.1906 & 0.1779 & \num{9.3778e-5} \\
        
        138 & 0.2223 & 0.1927 & 0.2080 & 0.0001 \\
        
        \bottomrule
        
    \end{tabular}
    \label{tab:drl-top-five-variances}
\end{table}

\begin{figure}
    \centering
    \subfloat[Untuned Model]{\includegraphics[scale=0.4]{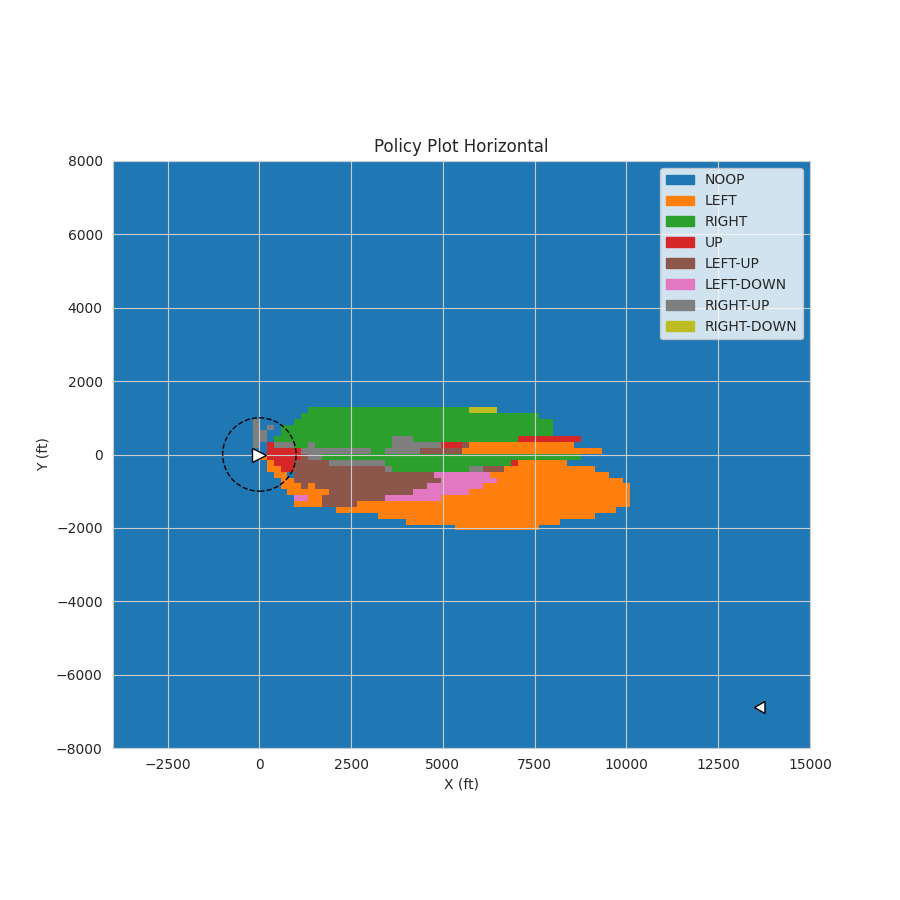}}
    \subfloat[Tuned Model]{\includegraphics[scale=0.4]{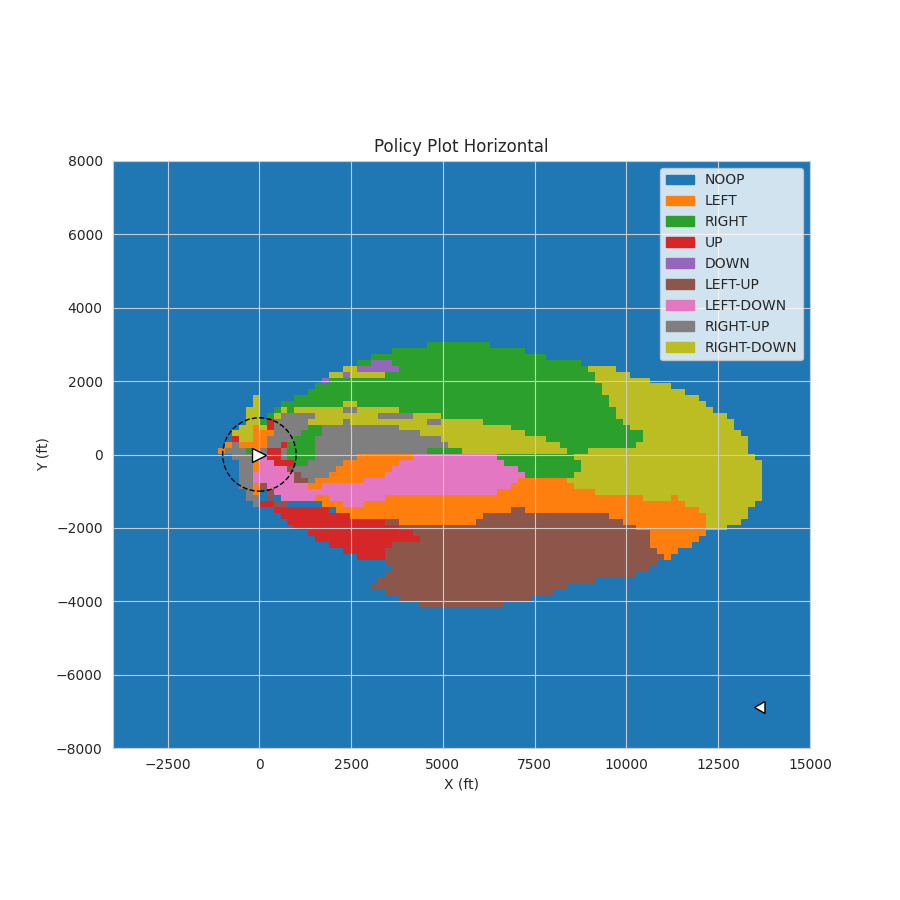}}
    \caption{Final DRL Model Policy Plots}
    \label{fig:drl-finalModelPolicyPlots}
\end{figure}

To compare our final results, we test the best model from iteration 46 against both the default reward model parameters and ACAS Xu. In section \ref{drl_performance}, we removed the DRL algorithm's lookahead capability to improve runtime performance. We reintroduce this capability to both DRL models and perform simulations with the same subset of the LLCEM encounters used to tune the reward model. The results from these simulations can be found in Table \ref{tab:drl-final-results}. As we can see from this table, we reduced the NMAC rate from the untuned version by 61\% while maintaining an alert rate only 11\% higher than the DP approach. Furthermore, our tuned model showed a reversal rate reduction of 75\%. The final policy plots for the tuned and untuned models can be found in Fig. \ref{fig:drl-finalModelPolicyPlots}. 

When examining the policy plot from our tuned model, we see that it is generally a sensible policy. It alerts in an oval-shaped region in front of the ownship aircraft and issues a Right maneuver if the intruder is to the left of the ownship aircraft's heading and a Left maneuver if the intruder is to the right of the aircraft's heading. One concern with the policy is that it will issue a combined maneuver before a single-dimension action in some situations. We can see that the ownship aircraft issues a combined Right-Down maneuver when the intruder is 10,000 feet directly in front of the intruder or a Left-Up maneuver if the intruder is towards the right edge of the ownship aircraft's alert region. Ideally, we would like to see a policy where the aircraft issues a single-dimension maneuver initially and then strengthens to a combined maneuver if the single-dimension maneuver fails to resolve the conflict.

When comparing this model to the untuned version, we can see that it issues an alert in a much larger region of space, causing the tuned alert rate to be about 35\% higher than the untuned model. The other main difference between the two policy plots is the amount of combined maneuvers issued from the tuned model. The policy plots show that the ratio of combined maneuvers to single-dimension maneuvers is higher in the tuned model than the untuned model. These two factors are the most likely cause for the increase in safety between the tuned and untuned models.

\begin{table}
    \centering
    \renewcommand{\arraystretch}{1.5}
    \caption{Final DRL Results}
    \begin{tabular}{clll}
        \toprule
        \textbf{Model} & \textbf{p(NMAC)} & \textbf{p(Alert)} & \textbf{p(Reversal)} \\
        \midrule
        
        DQN tuned & \num{2.4e-3} & 0.2164 & 0.0047 \\
        
        DQN untuned & \num{6.2e-3}  & 0.1604 & 0.0187 \\
        
        ACAS Xu & \num{9.8268e-4} & 0.1946 & 0.0029 \\
        
        \bottomrule
        
    \end{tabular}
    \label{tab:drl-final-results}
\end{table}

%% file: conclusion.tex
\section{Conclusion}

This project examined the application of a surrogate optimizer to the reward model for a deep reinforcement learning algorithm. With the surrogate optimizer, we were able to increase safety and improve operational capabilities while maintaining an acceptable alert rate. This improved system has reduced the performance gap between the DQN and DP algorithms.

Potential future work may include examining a variable training data NMAC rate. The training data for the DQN agent uses an NMAC rate elevated above the nominal rate. This appears to introduce a bias in the DQN agent to be overtly cautious, leading to many DQN agents that alert in all encounters regardless of the existence of NMAC. In addition, a different reward model could be structured to penalize combined maneuvers to favor initially alerting with single dimension maneuvers followed by multi-dimension maneuvers, as was preferred in the design of ACAS X variants.